\documentclass[letterpaper, 10 pt, conference]{ieeeconf}  

\IEEEoverridecommandlockouts                              

\usepackage{color}
\usepackage{caption}
\usepackage[textwidth=1.7cm]{todonotes}

\usepackage{graphics} 
\usepackage{amsmath} 
\usepackage{amssymb}  
\usepackage{biblatex} 
\usepackage{booktabs}
\usepackage{graphicx}
\usepackage[capitalise]{cleveref}

\usepackage{multirow}
\usepackage{amsmath,amsfonts,bm}
\usepackage{booktabs} 

\DeclareMathOperator*{\argmax}{arg\,max}

\DeclareMathOperator*{\argsortA}{arg\,sort}

\title{\LARGE \bf
We Need to Talk: Identifying and Overcoming\\\textit{Communication-Critical} Scenarios for Self-Driving
}

\author{Nathaniel Moore Glaser and Zsolt Kira\\
Georgia Institute of Technology\\
\texttt{$\{$nglaser,zkira$\}$@gatech.edu}
}
\begin{document}

\maketitle
\thispagestyle{empty}
\pagestyle{empty}

\begin{abstract}
In this work, we consider the task of collision-free trajectory planning for connected self-driving vehicles.  We specifically consider \textit{communication-critical} situations---situations where single-agent systems have blindspots that require multi-agent collaboration.  To \textit{identify} such situations, we propose a method which (1) simulates multi-agent perspectives from real self-driving datasets, (2) finds scenarios that are challenging for isolated agents, and (3) augments scenarios with adversarial obstructions.  To \textit{overcome} these challenges, we propose to extend costmap-based trajectory evaluation to a distributed multi-agent setting.  We demonstrate that our \textit{bandwidth-efficient}, \textit{uncertainty-aware} method reduces collision rates by up to $62.5\%$ compared to single agent baselines.
\end{abstract}

\section{Introduction}

\begin{figure}
\vspace{1mm}
\centering
\includegraphics[scale=0.32]{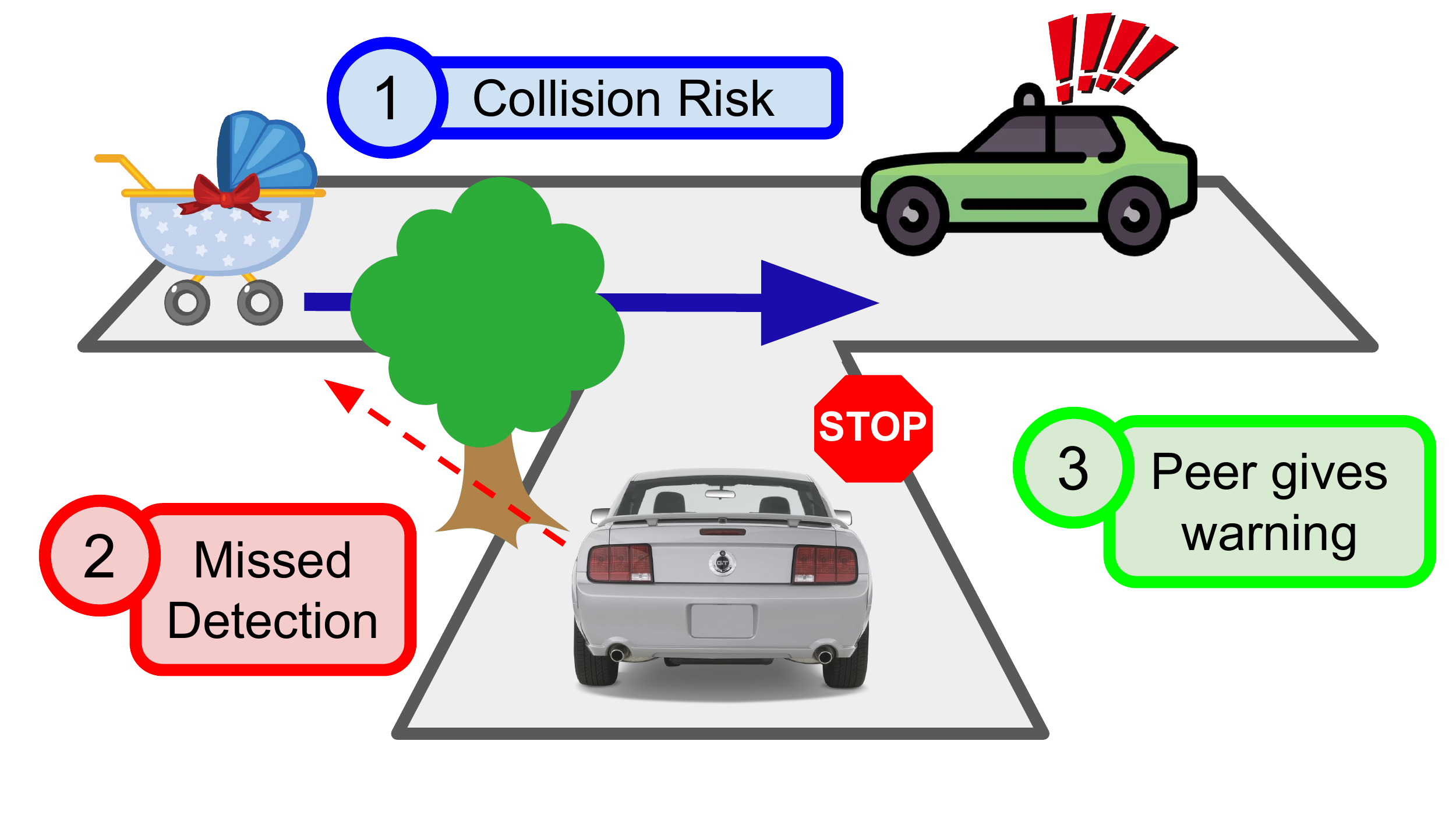}
\vspace*{-15pt}
\setlength{\belowcaptionskip}{-15pt}
\caption{\textbf{\textit{Communication-Critical} Planning.} In some cases, collaboration is critical for avoiding poor planning decisions.}
\label{fig:task}%
\end{figure}

Single agent systems are prone to perceptual degradations, such as the line-of-sight obstruction illustrated in Figure~\ref{fig:task}.  Such degradations are \textit{dangerous} and can potentially mislead downstream planning processes.  They are also \textit{dangerously rare}, existing in a ``long-tail'' regime where modern learning-based algorithms have historically performed poorly.  In this work, we explore how \textit{multi-agent} collaboration can provide a ``hedge'' against these rare-but-dangerous scenarios.  

Specifically, we address the setting of \textbf{\textit{communication-critical}} trajectory planning for connected self-driving vehicles.  In this setting, multi-robot communication is essential in order to reduce vehicle collision rates. 

First, we \textbf{\textit{identify}} these scenarios with a lightweight dataset wrapper, which ingests trajectory recordings from open-source datasets~\cite{interactiondataset} and synthesizes lidar scans.  We demonstrate an ability to \textit{{simulate}} and \textit{{identify}} interesting multi-agent scenarios as well as \textit{{augment}} less interesting ones.

Second, we \textbf{\textit{overcome}} these challenging scenarios with a novel method for performing distributed trajectory scoring.  Our method enables costmap-based planners to efficiently query the inferences of collaborators. We demonstrate that our method reduces hard collisions by up to $62.5\%$ compared to a single agent baseline. We demonstrate the benefits of several key components of our pipeline, which include (1) a learnable motion forecasting module that predicts future occupancy and perceptual uncertainty; (2) a bandwidth-efficient communicator selection mechanism; and (3) an uncertainty-aware mechanism for multi-agent score fusion.

\textbf{Related Work.}
Prior works in \textit{\textbf{collaborative perception}}, such as V2X-Sim~\cite{Li_2021_RAL}, use multi-agent collaboration to address failed inferences.  However, these failures often have limited impact on short-term planning, especially for far away inferences.  In response, our work searches for the subset of collaborative perception settings \textit{that also impact planning}.
To overcome these settings, we leverage prior work in \textit{\textbf{motion forecasting}} and \textit{\textbf{costmap-based planning}}.  Like Khurana \textit{et al.}~\cite{khurana2022differentiable}, our work uses a learned pipeline to forecast scene occupancy from top-down lidar rasters, and we similarly perform trajectory scoring based on these occupancy forecasts.  However, unlike their work, we extend this \textit{single agent} paradigm to a \textit{distributed multi-agent} one, leveraging efficient communication along the way~\cite{glaser2023communicationcritical}.  

\section{Methodology}

\subsection{\textbf{Identifying} Communication-Critical Scenarios}
\label{sec:dataset}

In this section, we propose a method to \textit{{simulate}}, \textit{{identify}}, and \textit{{augment}} scenarios that are interesting for connected self-driving vehicles.  Our method addresses two key challenges with real-world self-driving car datasets:

\begin{enumerate}
    \item \textbf{Single-Perspective}. Most existing SDV datasets are captured from \textit{single-agent} perspectives, and they do not naturally support multi-robot collaboration.  To overcome the lack of real, multi-robot SDV data, we use a lightweight simulator to synthesize the local perspectives of any number of actors in a shared scene.  
    \item \textbf{Rare-but-Dangerous}. Most scenarios for SDVs are straight-forward and may be resolved with \textit{single-agent} algorithms.  However, a small fraction of scenarios have severe implications (i.e. from missed detections); we want to assess performance in those critical situations.  To overcome the imbalance between common-but-easy and rare-but-dangerous scenarios, we propose to sift through real data to find the situations where communication is needed for safety.
\end{enumerate}

\textbf{Simulate}. To simulate multi-robot sensor information, we leverage the \textbf{Collaborative Bird's Eye View Simulator} (CoBEV-Sim) from prior work~\cite{glaser2023communicationcritical}.  This lightweight dataset wrapper ingests the motion tracks captured in overhead datasets~\cite{interactiondataset}, and it synthesizes local perception information (i.e. lidar scans) for any subset of those vehicles.  

\textbf{Identify}.  We identify communication-critical scenarios by inserting trajectories from the \textbf{Interaction} dataset~\cite{interactiondataset} into the \textbf{CoBEV-Sim} wrapper.  We subdivide this dataset into ``scenarios'' consisting of a $3$ second observation window and a $1$ second planning window.  Next, we randomly select a set of vehicles to be communication-enabled, and we generate a set of plausible future trajectory candidates for each.  Finally, we query the \textbf{CoBEV-Sim} wrapper for the agents that are \textit{invisible} during the observation window but \textit{collide} with those trajectory candidates during the planning window.  These collisions result from limited single-agent perception, which are rare and dangerous, and these scenarios will likely see the largest benefit from collaborative perception.  

\begin{figure}
\vspace{1mm}
\centering
\includegraphics[scale=0.55]{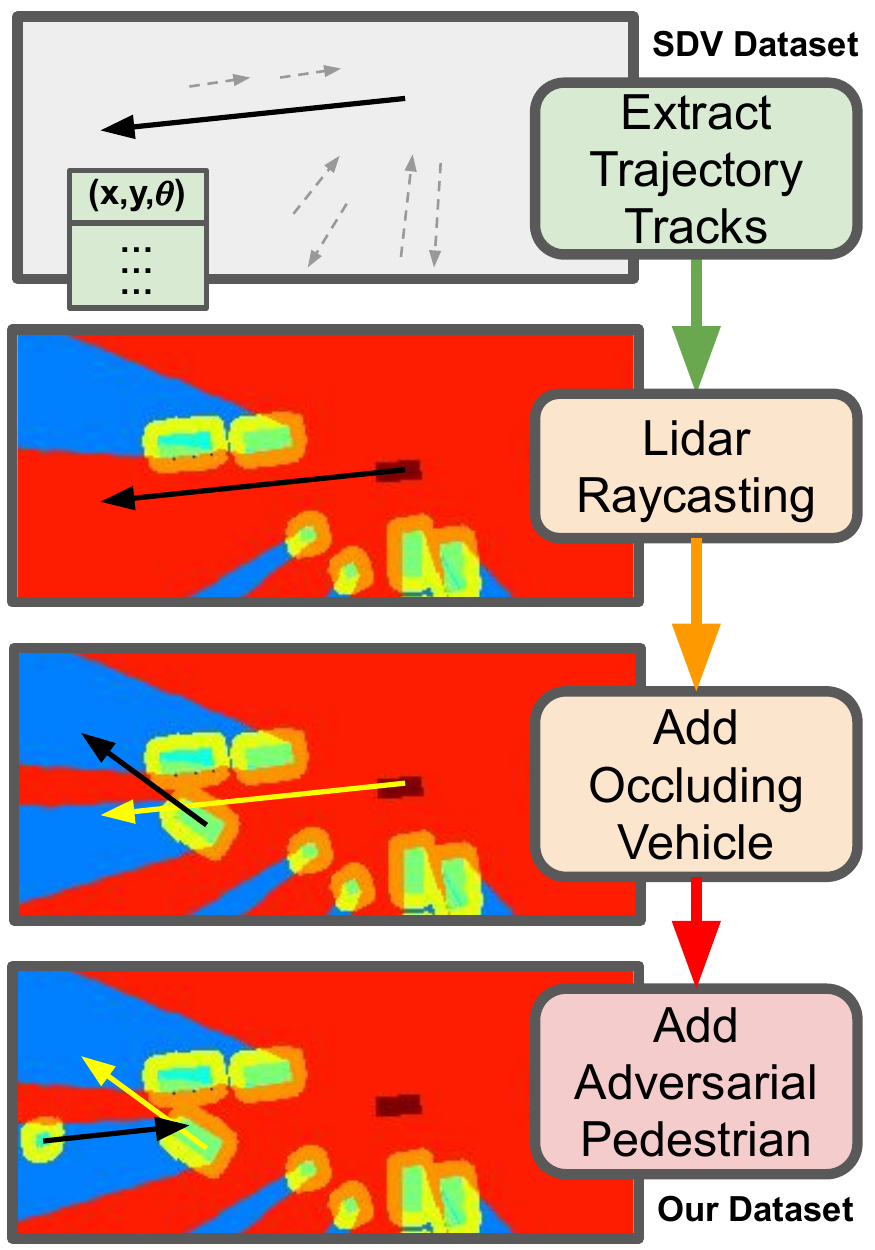}
\caption{\textbf{Adversarial Augmentations}. The trajectories from a seed dataset are rendered with lidar raycasting and augmented with occluding vehicles and adversarial pedestrians.
}
\label{fig:synthesize}%
\end{figure}

\textbf{Augment}. In response to the rarity of dangerous scenarios, we augment
remaining scenarios by introducing (1) an adversarial occluder and (2) an adversarial pedestrian, as summarized in Figure~\ref{fig:synthesize}.  The occluding agent is spawned such that it crosses in front of one of the communication-enabled agents, thereby blocking its sensors.  The adversarial pedestrian is spawned such that it travels in the ``shadow'' of this occluder during the observation window and intersects with the path of the agent during the planning window. 

\begin{figure}
\vspace{2mm}
\centering
\includegraphics[scale=0.6]{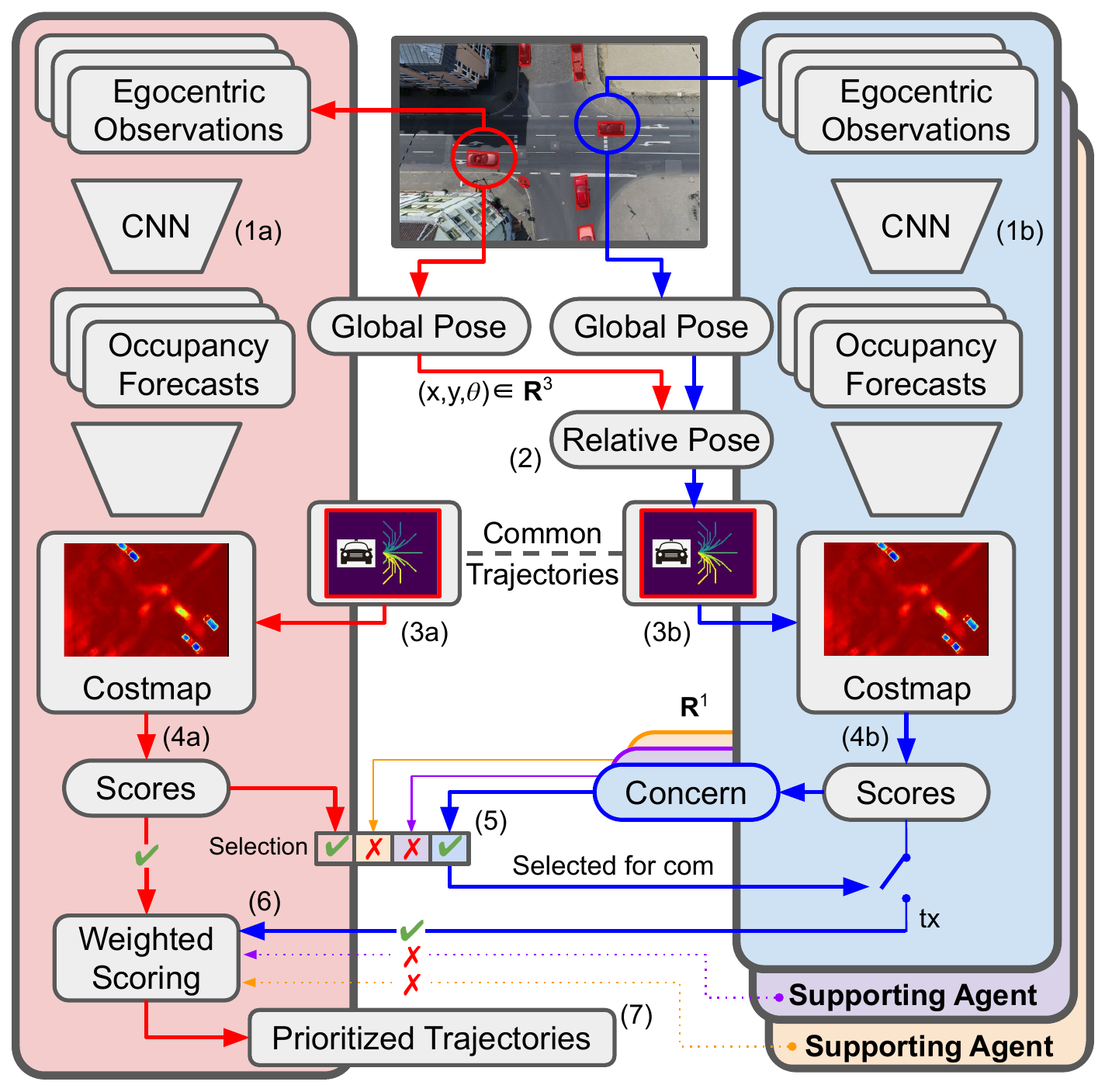}
\caption{\textbf{Distributed Trajectory Evaluation}.  The target agent (left: red) scores a set of trajectory candidates by leveraging the inferences of collaborators (right: blue, purple, yellow).  
}
\label{fig:architecture}%
\end{figure} 

\begin{figure*}
\vspace{1mm}
\centering
\includegraphics[scale=0.6]{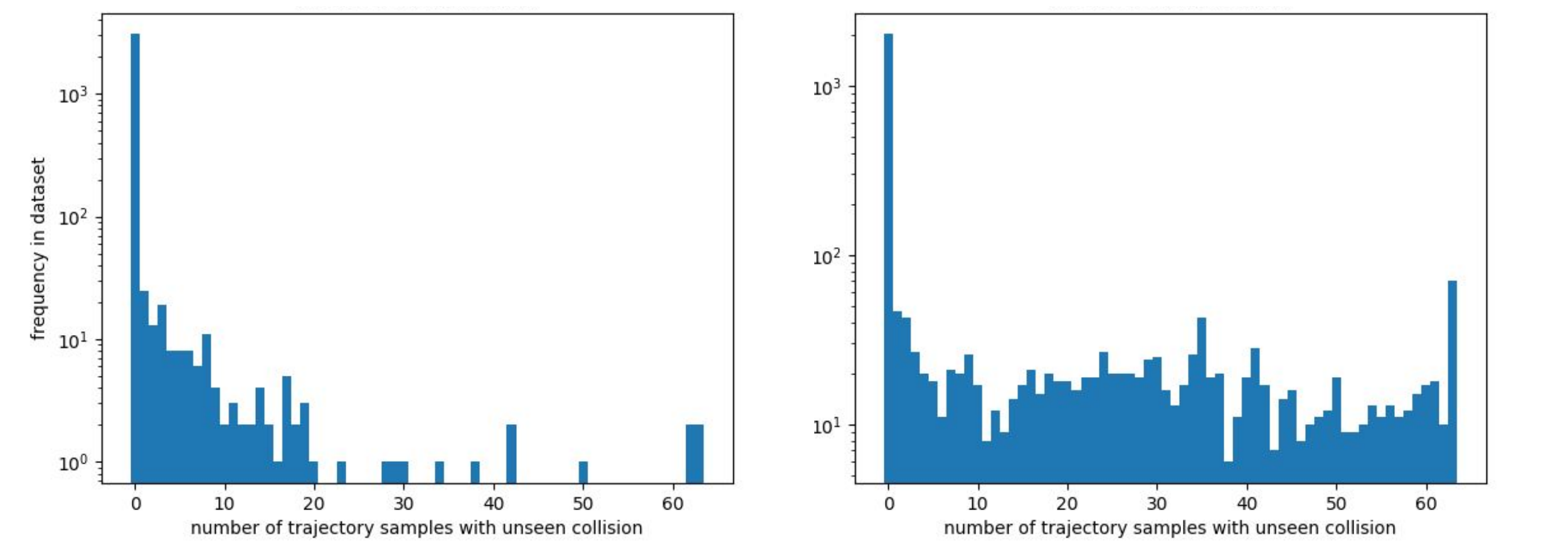}
\caption{\textbf{Prevalence of communication-critical scenarios before and after adversarial augmentations.}  We consider the number of trajectory candidates (out of $N=64$) that collide with \textit{unseen} road actors for each dataset scenario.  Datasets with more colliding trajectory candidates are rare but dangerous events that occur, which can be overcome via  \textit{collaboration with peers} to avoid making a dangerous planning decision.  Results are shown in log-scale for the \textbf{Interaction} dataset (left) and its adversarial augmentations (right).}
\label{fig:frequency}%
\end{figure*}

\subsection{\textbf{Overcoming} Communication-Critical Scenarios}
\label{sec:method}

In this section, we propose a method for \textit{overcoming} the perceptual adversity posed by situations identified in the previous section.  Specifically, our method allows for costmap-based planners to leverage distributed inferences, as shown in Figure~\ref{fig:architecture}.

\textbf{Motion Forecasting}. First, each agent uses a CNN $\bm{\Phi}$ to convert a sequence of aligned 2D Bird's Eye View (BEV) observations $\bm{X}^{T^-:0}$ into a sequence of 2D motion forecasts: 
\begin{equation}\label{eq:modelprediction}
    \bm{P}_{C}^{1:T^+} = \bm{\Phi}(\bm{X}^{T^-:0})\\
\end{equation}
\begin{equation}\label{eq:classes}
    C = \{\textbf{empty}, \textbf{occupied}, \textbf{shadow}, \textbf{outOfRange}\}
\end{equation}
where $\bm{P}_C^{1:T^+}$ refers to the softmax-normalized temporal confidence maps.  These confidence maps are trained via cross-entropy loss against the ground truth future BEV observations.  A channel-wise $\argmax$ yields semantic masks:
\begin{equation}\label{eq:semanticsequence}
  \bm{M}^{1:T^+} = \argmax_{C} \bm{P}^{1:T^+}
\end{equation}
Note that each agent \textit{explicitly} predicts perceptual uncertainty (with the \textbf{shadow} class) in addition to future occupancy (with the \textbf{occupied} class).  These outputs let each agent articulate when it is uncertain about its predictions, allowing for confident agents to assert themselves over uncertain ones.  

\textbf{Costmap Generation}. Next, each agent converts its 2D semantic forecasts into a costmap (Fig.~\ref{fig:architecture}.1).  Namely, each agent computes a signed distance field (\textbf{SDF})~\cite{oleynikova2016signed} on a binary mask sequence from the \textbf{occupied} class only:
\begin{equation}\label{eq:sdf}
  \bm{D}^t = \textbf{SDF}(\bm{M}^t) \text{ where } \bm{M} = \textbf{occupied}
\end{equation}

\textbf{Trajectory Generation}.  Each agent maintains an identical set of $N$ local trajectory candidates, $\mathbb{S} = \{S_{i}\}^N$, where each trajectory candidate defines a sequence of agent positions and headings at future timesteps,  $S_{i} = \{(x_t,y_t,\theta_t)\}_1^{T^+}$.  The trajectories were generated to cover a wide range of feasible vehicle maneuvers, as dictated by linear and angular acceleration limits (Fig.~\ref{fig:architecture}.3).

\textbf{Distributed Costmap Extraction}. In this section, we describe the procedure for extracting costmap values from supporting agent $a$ for trajectory candidate $i$.  First, each supporting agent receives a broadcasted target pose and transforms its static set of local trajectories to the location of the target agent (Fig.~\ref{fig:architecture}.2):
\begin{equation}\label{eq:trajectorysampling}
  S^t_{a,i} = R_a^e S^t_{e,i}
\end{equation}
where $R_a^e$ refers to the relative pose transform between the egocentric agent $e$ and supporting agent $a$. 

Next, each agent computes a trajectory score by extracting values at trajectory waypoint ($S^t$) from the \textbf{SDF} ($\bm{D}^t$) (Fig.~\ref{fig:architecture}.4):
\begin{equation}\label{eq:trajectorysampling}
    C^{t} = \bm{D}^{t}[S^t]_{\text{min}}
\end{equation}
where $[\cdot]_{\{\text{max},\text{min},\text{avg}\}}$ denotes the costmap extraction strategy at a specific trajectory pose.  In this work, we project the vehicle footprint onto the designated pose and compute a maximum (\textbf{max}), minimum (\textbf{min}), or average (\textbf{avg}) across all of the pixels that fall under that footprint.

Moreover, to assist with uncertainty-aware fusion, each agent extracts several other quantities from its forecasts:
\begin{equation}
    f_o = \max_t{(\bm{M}^{t}[S^t]_{\text{avg}})} \text{ where }  \bm{M} = \textbf{occupied}
\end{equation}
\begin{equation}
    p_o = \sum_t{\bm{P}_C^t[S^t]_{\text{avg}}} \text{ where }  C = \textbf{occupied}
\end{equation}
\begin{equation}
    f_s = \sum_t{(\bm{M}^{t}[S^t]_{\text{avg}})} \text{ where }  \bm{M} = \textbf{shadow}
\end{equation}
\begin{equation}
    p_s = \sum_t{\bm{P}_C^t[S^t]_{\text{avg}}} \text{ where }  C = \textbf{shadow}
\end{equation}

\textbf{\textit{Bandwidth-Efficient} Agent Selection}. To avoid saturating the communication channel, each supporting agent computes a single scalar value that reflects how ``concerned'' it is about the trajectory plans for the ego agent (Fig.~\ref{fig:architecture}.5):
\begin{equation}
    w = \sum_i{p_o^i}
\end{equation}
where $p_o^i$ refers to the total accumulated confidences for the \textbf{occupied} class, for trajectory $i$.  Next, the egocentric agent compares the scores from itself and its supporting agents to determine candidates for further communication:
\begin{equation}
    \overline{a} = \argmax_{a\in\text{all}}{w_a}
\end{equation}

\textbf{Score Fusion (\textit{Naive})}. At this point, each agent has computed the score of each trajectory candidate ($i$) at each timestamp ($t$).  These scores may be (naively) combined together to produce a single score per trajectory:
\begin{equation}
  F_i^{\text{all}} = \sum_{a \in \text{all}} \min_t{C_{a,i}^{t}}
\end{equation}

\textbf{Score Fusion (\textit{Selective})}.  To conserve bandwidth, we first propose communicating with the ``most concerned'' agent:
\begin{equation}
  F_i^{\text{concern}} = \sum_{a \in \{\text{ego,$\overline{a}$}\} } \min_t{C_{a,i}^{t}}
\end{equation}

\textbf{Score Fusion (\textit{Uncertainty-Aware})}.  To allow confident agents to assert themselves over less confident ones, we further propose uncertainty-weighted fusion (Fig.~\ref{fig:architecture}.6):
\begin{equation}
  F_i = \sum_{a \in \{\text{ego,$\overline{a}$}\} } u_{a,i} \min_t{C_{a,i}^{t}}
\end{equation}
\begin{equation}\label{eq:uncertainty}
   u = \frac{(1 + f_o) (1 + p_o)}{(1 + f_s) (1 + p_s)}
\end{equation}
where the uncertainty weighted term ($u$) is proportional to occupancy ($f_o, p_o$) and inversely proportional to perceptual uncertainty ($f_s, p_s$).  This term assigns more weight to trajectories that are predicted to collide (i.e. based on overlap with the \textbf{occupied} binary mask and confidence map) and down-weights trajectories that have significant perceptual uncertainty (i.e. based on overlap with the \textbf{shadow} class).

\textbf{Trajectory Prioritization}.  Finally, the set of trajectories is sorted based on their fused scores (Fig.~\ref{fig:architecture}.7):
\begin{equation}\label{eq:multiagentfusion}
    \mathbb{S}_{\text{prioritized}} = [S_i | \forall i \in \argsortA_i F_i]
\end{equation}







    
\begin{table*}
\vspace{5mm}
\centering
\resizebox{0.85\linewidth}{!}{%
\begin{tabular}{lcc||cc|cc||c}
                            &                           & Available &           \multicolumn{4}{c||}{Top-$k$ Collision Rate (\%)}                                               & Avg Num \\
                           & Method                     & $N_{com}$                &              $k=1$ &                  (rel to ego) &             $k=10$ &                  (rel to ego) & Com Links\\
\midrule
\midrule
Baselines                  & ego                        & 0               &              5.8 &                        100 &              14.3 &                        100 &       0.000\\
                           & randTraj                   & 0               &              48.6 &                       837 &              52.2 &                        365.5 &       0.000\\
                           & rand                       & 1               &              8.0 &                        137 &              22.3 &                        156.3 &       0.783\\
                           & ego + all                    & 3               &              4.3 &                        75.0 &              14.0 &                        98.0 &       2.862\\
\cmidrule(lr){2-8}                           
Ground Truth Vision        & \color{gray}ego*                       & \color{gray}0                &              \color{gray}1.4 &                        \color{gray}25.0 &              \color{gray}12.5 &                        \color{gray}87.8 &       \color{gray}0.000\\
\midrule
\midrule
Contribution A:            & ego + concern                & 1               &              4.3 &                        75.0 &              8.7 &                        60.9 &       0.754\\
(com with concerned)       &                            & 2               &              \textbf{3.6} &                        \textbf{62.5} &              \textbf{7.8} &                        \textbf{54.8} &       1.348\\
                           &                            & 3               &              4.3 &                        75.0 &              11.0 &                        77.2 &       2.080\\
\midrule
\midrule
Contribution A+B:          & ego + concern                & 1               &              3.6 &                        62.5 &              7.8 &                        54.8 &       0.754\\
(fuse with uncertainty)    &                            & 2               &              2.9 &                        50.0 &              6.6 &                        46.2 &       1.348\\
                           &                            & 3               &              \textbf{2.2} &                        \textbf{37.5} &              \textbf{5.5} &                        \textbf{38.6} &       2.080\\

\end{tabular}
}
\setlength{\belowcaptionskip}{-15pt}
\caption{\textbf{Top-$k$ collision rate with $n$-agent communication}.  This table shows the collision rates of our method compared to baselines.  The (rel to ego) column shows the percent change in collision rate as compared to a single agent (first row).}
\label{tab:collisionRateTable}
\end{table*}

\section{Results}

We assess performance by analyzing \textit{collision rates}.  To \textbf{identify} communication-critical scenarios, we compute the collision rate of trajectory candidates with unseen obstacles.  To \textbf{overcome} these challenging scenarios, we show reduced collision rates using a bandwidth-efficient, uncertainty-aware method for distributed multi-agent trajectory scoring.

\subsection{\textbf{Identifying} Communication-Critical Scenarios}
In Figure~\ref{fig:frequency}, as shown by the large vertical bars at $x=0$, most scenarios for self-driving vehicles have sufficient single-agent perception information for safe planning.  In these figures, $x=0$ refers to the fact that $0$ trajectory candidates are at risk of colliding with an unseen actor. However, as shown by the bars at $x>0$, a rare (but substantial) number of scenarios involve planning where single-agent perception is ``blind'' to some collision possibilities.

To better inoculate algorithms against these rare events, we augmented the dataset with adversarially placed scene actors. Our augmentations increase the number of interesting scenarios for multi-robot planning, as shown by the increase in mass at $x>0$ between the left and right plots.

\subsection{\textbf{Overcoming} Communication-Critical Scenarios}

As shown in Table~\ref{tab:collisionRateTable}, our method significantly reduces collision rates compared to single-agent and multi-agent baselines; and it approaches the performance of an egocentric agent that has omniscient, perfect vision.  We also demonstrate that our method has the added benefit of \textit{choice}, where a trajectory from the top $10$ candidates ($k=10$) may be selected with only slight performance loss.  Moreover, we highlight that our bandwidth-efficient communicator selection mechanism (Contribution A) and our uncertainty-aware multi-agent score fusion (Contribution B) both combine to reduce collision rates to $37.5\%$ of the egocentric agent, while also reducing the amount of communication connections required for different numbers of available communicators ($N_{com}$): $1 \rightarrow 0.75$, $2 \rightarrow 1.35$, $3 \rightarrow 2.08$.
\section{Acknowledgement}
\label{sec:acknowledgement}
\noindent This work was supported by ONR grant N00014-18-1-2829.

\section{Conclusion}

In this work, we propose a method to \textbf{identify} and \textbf{overcome} communication-critical scenarios.  We demonstrate that we are able to (1) \textit{simulate} multi-agent perspectives from single-agent datasets, (2) \textit{identify} scenarios that are interesting for multi-agent collaboration, and (3) \textit{augment} less interesting scenarios---all to better test collaborative perception and planning algorithms.  Moreover, we demonstrate that our bandwidth-efficient, uncertainty-aware method reduces collision rates to $37.5\%$ of single-agent performance in these challenging situations.

\textbf{Future Work}. Future work will explore the challenges that arise from considering more realistic multi-agent components, such as noisy sensors, imprecise pose estimates, and communication latency.

\printbibliography

@article{interactiondataset,
  title={Interaction dataset: An international, adversarial and cooperative motion dataset in interactive driving scenarios with semantic maps},
  author={Zhan, Wei and Sun, Liting and Wang, Di and Shi, Haojie and Clausse, Aubrey and Naumann, Maximilian and Kummerle, Julius and Konigshof, Hendrik and Stiller, Christoph and de La Fortelle, Arnaud and others},
  journal={arXiv preprint arXiv:1910.03088},
  year={2019}
}

@ARTICLE{Li_2021_RAL,
  author={Li, Yiming and Ma, Dekun and An, Ziyan and Wang, Zixun and Zhong, Yiqi and Chen, Siheng and Feng, Chen},
  journal={IEEE Robotics and Automation Letters}, 
  title={V2X-Sim: Multi-Agent Collaborative Perception Dataset and Benchmark for Autonomous Driving}, 
  year={2022},
  volume={7},
  number={4},
  pages={10914-10921},
  doi={10.1109/LRA.2022.3192802}}

@article{glaser2023communicationcritical,
  title={Communication-Critical Planning via Multi-Agent Trajectory Exchange},
  author={Glaser, Nathaniel Moore and Kira, Zsolt},
  journal={arXiv preprint arXiv:2303.06080},
  year={2023}
}

@inproceedings{oleynikova2016signed,
  title={Signed distance fields: A natural representation for both mapping and planning},
  author={Oleynikova, Helen and Millane, Alexander and Taylor, Zachary and Galceran, Enric and Nieto, Juan and Siegwart, Roland},
  booktitle={RSS 2016 Workshop: Geometry and Beyond-Representations, Physics, and Scene Understanding for Robotics},
  year={2016},
  organization={University of Michigan}
}

@inproceedings{khurana2022differentiable,
  title={Differentiable raycasting for self-supervised occupancy forecasting},
  author={Khurana, Tarasha and Hu, Peiyun and Dave, Achal and Ziglar, Jason and Held, David and Ramanan, Deva},
  booktitle={Computer Vision--ECCV 2022: 17th European Conference, Tel Aviv, Israel, October 23--27, 2022, Proceedings, Part XXXVIII},
  pages={353--369},
  year={2022},
  organization={Springer}
}
\end{document}